\documentclass{svproc}
\usepackage{url,graphicx}
\usepackage{HamidPackage}

\begin{document}
\mainmatter              % start of a contribution
%
% \title{Hamiltonian Mechanics unter besonderer Ber\"ucksichtigung der
% h\"ohereren Lehranstalten}
%
\title{New Variants of Frank-Wolfe Algorithm for Video Co-localization Problem}
% \titlerunning{Hamiltonian Mechanics}
\titlerunning{FW Variants in Video Co-Localization}  % abbreviated title (for running head)
\author{Hamid Nazari}
%
% \authorrunning{Ivar Ekeland et al.} % abbreviated author list (for running head)
% \tocauthor{Ivar Ekeland, Roger Temam, Jeffrey Dean, David Grove,
% Craig Chambers, Kim B. Bruce, and Elisa Bertino}
%
\institute{Clemson University, Clemson, SC}

\maketitle              % typeset the title of the contribution

\begin{abstract}

The co-localization problem is a model that simultaneously localizes objects of the same class within a series of images or videos. In \cite{joulin2014efficient}, authors present new variants of the Frank-Wolfe algorithm (aka conditional gradient) that increase the efficiency in solving the image and video co-localization problems.  The authors show the efficiency of their methods with the rate of decrease in a value called the Wolfe gap in each iteration of the algorithm. In this project, inspired by the conditional gradient sliding algorithm (CGS) \cite{CGS:Lan}, We propose algorithms for solving such problems and demonstrate the efficiency of the proposed algorithms through numerical experiments. The efficiency of these methods with respect to the Wolfe gap is compared with implementing them on the YouTube-Objects dataset for videos.
% The abstract should summarize the contents of the paper
% using at least 150 and at most 200 words. It will be set in 9-point
% font size and be inset 1.0 cm from the right and left margins.
% There will be two blank lines before and after the Abstract. \dots
% We would like to encourage you to list your keywords within
% the abstract section using the \keywords{...} command.
\end{abstract}
\section{Image and Video Co-Localization Problems}
Problems in recognizing and localizing particular objects in images and videos have received much attention recently as internet photo and video sharing have become increasingly popular. 
Co-localization involves localizing with bounding boxes in a set of images or videos as a sequence of images (frames).

\section{Model Setup for Images}

Our ultimate goal is to localize the common object in a set of images or in a series of frames of a video. Here we first have a brief review of image and video models based on formulation in \cite{joulin2014efficient}. To this end we review the required back grounds in each step as much as the features and variables in the mathematical programming model become understandable. Note that this formulation is based on formulation introduced in \cite{tang2014co} for image co-localization. Quadratic formulation that we review in this section localizes any set of images and videos, simultaneously. In \cite{boykov2001fast,delong2012minimizing,delong2012fast} also, we can find similar discrete optimization approaches in various computer vision applications. 

\subsection{Objectness for Images}

Suppose that we have a set $\mathcal{I} = \{I_1, I_2, \dots, I_n\}$ of $n$ given images, and our goal is to localize the common object in each image. One approach is to find candidate boxes in each image that potentially contain an object using \emph{objectness} \cite{alexe2012measuring}. 

\begin{figure}[!ht]
	\centering
	\includegraphics[scale=.3]{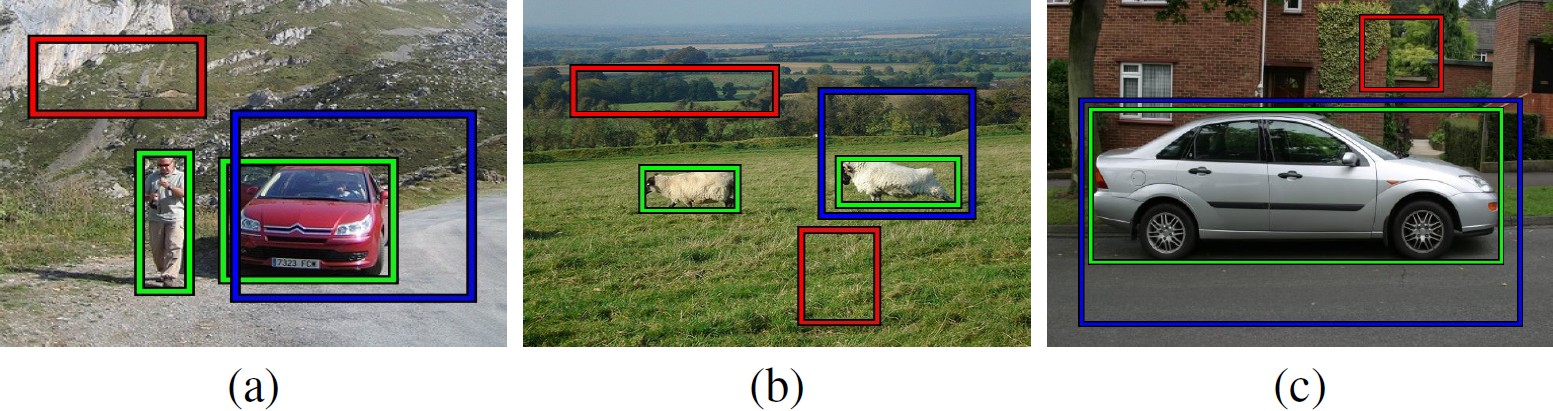}
	\caption{The objectness measure should score the blue windows, partially covering the objects, lower than the ground truth windows in green, and score even lower the red windows containing only the background. Image is from \cite{alexe2012measuring}.}
	\label{fig:localization:objectness2}
\end{figure}

While object detectors for images are usually specialized for one object class such as cars, airplanes, cats, or dogs, objectness quantifies how likely it is for an image window to cover an object of any class. In an image, objects have a well-defined boundary and center, cats, dogs, and chairs, as opposed to indefinite background, such as walls, sky, grass, and road. Figure \ref{fig:localization:objectness2} illustrates the desired behavior of an objectness measure. Green windows must score highest windows fitting an object tight, blue windows should score lower windows covering partly an object and partly the background, and red windows should score lowest windows containing only partial background. This approach and the way we score the windows is designed in \cite{alexe2012measuring} and explicitly trained to distinguish windows containing an object from background windows. 

Using objectness, we generate $m$ candidate boxes (e.g. green boxes in Figure \ref{fig:localization:objectness2}) for each image that could potentially contain an object. In other words, if $j\in \{1,2,\dots,n\}$ we define $\mathcal{B}_j$ to be the set of all boxed in image $I_j\in\mathcal{I}$. Then the goal is to select the box that contains the object, from each image, jointly. Also. for simplicity let $\mathcal{B} = \mathcal{B}_1 \cup \mathcal{B}_2 \cup \cdots\cup \mathcal{B}_n$ and $n_b = nm$ the total number of boxes in all images.

\subsection{Feature representation}

Assume that we have determined $m$ candidate boxes in each of two the different images $I_i$ and $I_j$ for any $i,j\in\{1,2,\dots, m\}$. A common object in $I_i$ and $I_j$ might be in different shape, scale, color, brightness, angle and many other features. Therefore, it is critical to extract distinctive invariant features from images that can be used to perform reliable matching between different views of an object. David G. Lowe in \cite{lowe2004distinctive} introduces a method that finds features that are invariant to image scaling and rotation, and partially invariant to change in illumination and 3D camera view point. Using his method, large number of features can be extracted from typical images with efficient algorithms, as well as the cost of extracting these features is minimized. The major stages of computation used to generate the set of image features are as follows. 
\begin{enumerate}[1.]
\item \textbf{Scale-space extrema detection:} The first stage of computation searches over all scales and image locations. It is implemented efficiently by using a difference-of-Gaussian function to identify potential interest points that are invariant to scale and orientation.

\item \textbf{Keypoint localization:}
At each candidate location, a detailed model is fit to determine location and scale. Keypoints are selected based on measures of their stability.

\item \textbf{Orientation assignment:}
One or more orientations are assigned to each keypoint location based on local image gradient directions. All future operations are performed on image data that has been transformed relative to the assigned orientation, scale, and location for each feature, thereby providing invariance to these transformations.

\item \textbf{Keypoint descriptor:}
The local image gradients are measured at the selected scale in the region around each keypoint. These are transformed into a representation that allows for significant levels of local shape distortion and change in illumination.

\end{enumerate}
This process is called Scale Invariant Feature Transform (SIFT). SIFT transforms image data into scale-invariant coordinates relative to local features. Using SIFT we can generate large numbers of features that densely cover the image over full range of scales and locations.

Let $b_k$ be a box in $\mathcal{B}$. Then we denote the SIFT feature representation of $b_k$ as $x_k\in \R^d$ where $d = 10,000$ is the dimensional feature descriptor for each box in $\mathcal{B}$. Finally, we stack the feature vectors to form a feature matrix $X\in \R^{n_b\times d}$.

\subsection{Prior, Similarity, and Discriminability of boxes}

Let us denote the boxes that contain an instance of the common object as \emph{positive} boxes, and the ones that don't as \emph{negative} boxes. Then a prior is introduced for each box that represents a score that the box is positive. This happens using a saliency map \cite{perazzi2012saliency} for each box and the prior is in fact the average saliency within the box, weighted by the size of the box. Finally we stack these values into the $n_b$ dimensional vector $\vec{m}$ as the prior vector.

\begin{figure}[!ht]
	\centering
	\includegraphics[scale=.25]{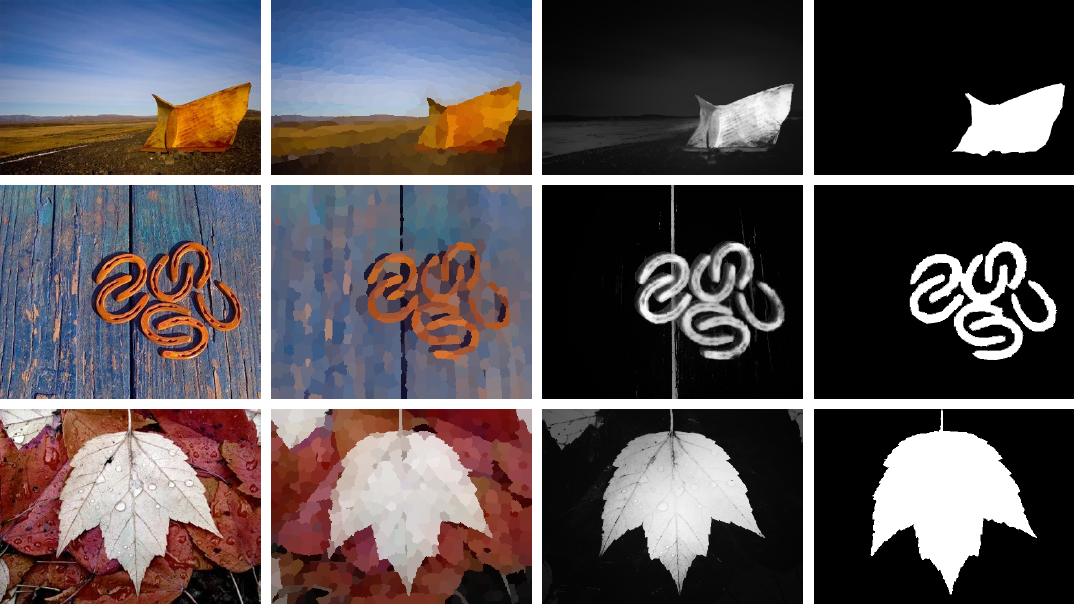}
	\caption{ An example of saliency mappings for images from left to right. Image is from [41].}
	\label{fig:localization:saliency1}
\end{figure}

In addition, boxes that have the similar appearance should be labeled the same. This happens through a matrix called similarity matrix denoted by $S$. Similarity matrix of boxes in $\mathcal{B}$ is based on the box feature matrix $X$ described above. Let $b_i$ and $b_j$ be any two boxes in $\mathcal{B}$ where $i,j\in\{1,2,\dots,n_b\}$. Then similarity matrix $S\in\R^{n_b\times n_b}$ is computed based on the $\chi^2$-distance as
\begin{align}\label{local:similarity_matrix}
	S_{ij} = \exp\cbr{-\gamma\sum_{k=1}^{d}\frac{(x_{ik} - x_{jk})^2}{x_{ik} + x_{jk}}},
\end{align}
where $\gamma = (10d)^{-1/2}$. For $i$ and $j$ where boxes $b_i$ and $b_j$ belong to the same image we set $S_{ij}=0$. Then the normalized Laplacian matrix \cite{shi2000normalized} is computed as
\begin{align}\label{localization:laplacian}
	\mathcal{L} = I_{n_b} - D^{-1/2}SD^{-1/2},
\end{align}
where $D$ is the diagonal matrix composed of row sums of $S$.

\subsection{Model Formulation}

Associated with each box $b_{j,k}\in\mathcal{B}_j$ we define a binary variable $z_{j,k}$ where $z_{j,k}=1$ when $b_{j,k}$ is a positive box (contains an instance of the common object) and 0 otherwise. Then we define the integer vector variable 
\begin{align}\label{localization:formulation:variable}
\vec{z} = (z_{1,1},\dots,z_{1,m}, \dots, z_{n,1},\dots, z_{n,m})^T\in\{0,1\}^{n_b}. 
\end{align}
Making the assumption that in each image there exist at most 1 positive box, our set of constraints are define by
\begin{align}\label{localization:formulation:const1}
	\sum_{k = 1}^{m} z_{j,k} = 1,\qquad \forall j \in \{1,\dots, n\}.
\end{align}

As we introduced a prior for each box and defined the $n_b$ dimensional vector of average saliency within the boxes, we obtain a linear term that penalizes less salient boxes as part of the objective function:
\begin{align}\label{localization:formulation:obj1}
	f_p(\vec{z}) := -\vec{z}^T\log(\vec{m}).
\end{align}
Similarly, our choice of normalized Laplacian matrix $\mathcal{L}$ defined in \eqref{localization:laplacian} results in a quadratic term that handles the selection of similar boxes:
\begin{align}\label{localization:formulation:obj2}
	f_L(\vec{z}) := \vec{z}^T\mathcal{L}\vec{z}.
\end{align}
This is motivated by the work of Shi and Malik \cite{shi2000normalized} in which they have taken advantage of eigenvalues of the Laplacian for clustering $\vec{z}$ by the similarity matrix. In fact, they have shown that with the eigenvector corresponding to the second smallest eigenvalue of a normalized Laplacian matrix we can cluster $\vec{z}$ along the graph defined by the similarity matrix, leading to normalized cuts when used for image segmentation. Also, Belkin and Niyogi \cite{belkin2003laplacian} showed that this problem is equivalent to minimizing \eqref{localization:formulation:obj2} under linear constraints. In fact, the similarity term works as a generative term which selects boxes that cluster well together  \cite{tang2014co}.

Although discriminative learning techniques such as support vector machines and ridge regression has been widely used on many supervised problems in which there are know labels, they can be used in this unsupervised case where the labels of boxes are unknown \cite{bach2007diffrac,xu2004maximum}. Motivated by \cite{joulin2010discriminative}, we consider the ridge regression objective function for boxes:
\begin{align}\label{localization:formulation:ridge}
	\min_{w\in\R^d,\ c\in\R}\quad \frac{1}{n_b}\sum_{j=1}^{n}\sum_{k=1}^{m}\norm{z_{j,k}-wx_{j,k} - c}_2^2 - \frac{\kappa}{d}\norm{w}_2^2,
\end{align}
where $w$ is the $d$ dimensional weight vector of the classifier, and $c$ is the bias. This cost function is being used among discriminative cost functions because the ridge regression problem has a explicit (closed form) solution for weights $w$ and bias $c$ which implies the quadratic function in the box labels \cite{bach2007diffrac}:
\begin{align}\label{localization:formulation:obj3}
	f_D(\vec{z}):=\vec{z}^T\mathcal{A}\vec{z},
\end{align}
where
\begin{align}\label{localization:formulation:A}
	\mathcal{A}= \frac{1}{n_b}\Pi_{n_b}\pr{I_{n_b}-X(X^T\Pi_{n_b}X+n_b\kappa I_{n_b})^{-1}X^T}\Pi_{n_b},
\end{align}
is the discriminative clustering term and $	\Pi_{n_b} = I_{nb} - \frac{1}{n_b}\vec{1}_{n_b}\vec{1}_{n_b}^T$ in \eqref{localization:formulation:A} is the centering projection matrix. Note that this quadratic term allows us to utilize a discriminative objective function to penalize the selection of boxes whose features are not easily linearly separable from other boxes.

Summing up our results in \eqref{localization:formulation:const1}, \eqref{localization:formulation:obj1}, \eqref{localization:formulation:obj2}, and \eqref{localization:formulation:obj3}, the optimization problem to select the best box in each image is given by
\begin{align}\label{localization:formualtion:img_problem}
	\begin{aligned}
		\min_{\vec{z}}&\qquad \vec{z}^T(\mathcal{L}+\mu\mathcal{A})\vec{z} - \lambda\ \vec{z}^T\log(\vec{m})\\
		\text{s.t}    &\qquad \sum_{k = 1}^{m} z_{j,k} = 1,\qquad j=1,\dots, n\\
			           &\qquad \vec{z} = (z_{1,1},\dots,z_{1,m}, \dots, z_{n,1},\dots, z_{n,m})^T\in\{0,1\}^{n_b},
	\end{aligned}
\end{align}
where parameter $\mu$ regularizes the trade-off between the quadratic terms \eqref{localization:formulation:obj2} and \eqref{localization:formulation:obj3}, and parameter $\lambda$ handles the trade-off between the linear term \eqref{localization:formulation:obj1} and the quadratic terms \eqref{localization:formulation:obj2} and \eqref{localization:formulation:obj3}. Recall that the linear constraints ensures that one box from each image is selected in the optimal solution. Note that Hastie, Tibshirani, and Friedman in \cite{hastie2009elements} showed that $\mathcal{A}$ is a positive semi-definite matrix. Also, since matrix $\mathcal{L}$ is positive semi-definite as well, the objective function of \eqref{localization:formualtion:img_problem} is convex. 

\section{Model Setup for Videos}
Co-localization in a video is very similar to the image case, as a video is a sequence of images that are called frames. While an object might not have an extreme change in size, shape, color, etc in two frames in row, co-localization in a video could be a simpler task at some point. In this section we describe the localization of a common object in a set of videos. In fact, if $\mathcal{V} = \{V_1, V_2, \dots, V_n\}$ is a set of $n$ given videos, we explore an approach to localize a common object in each frame of each video. More precisely, we consider $\mathcal{I}_i = \{I_{i1}, I_{i2}, \dots, I_{il_i}\}$ to be the temporally ordered set of frames of video $V_i$. Here $I_{ij}$ is the $i$-th frame of the $j$-th video and $l_i$ is the total number of frames, or the length of $V_i$ for $i=1,\dots,n$ and $j=1,\dots, l_i$. Similar to what we did in image case, we set $\mathcal{B}_{i,j}$ to be the set of $m$ generated candidate boxes, using objectness \cite{alexe2012measuring}, for $j$-th of $i$-th video. Then, considering $l_i$ frames in video $i$ and m boxes in each frame, we set $n_b^v = \sum_{i=1}^{n} l_im$ to be the total number of boxes in $\mathcal{V}$, the set of all videos. 

Note that, if we set $\mathcal{I} = \{\mathcal{I}_1, \mathcal{I}_2,\dots, \mathcal{I}_n\}$ to be the ordered set of all frames in $\mathcal{V}$, model \eqref{localization:formualtion:img_problem} returns a single box in each frame (image) as an optimal solution. Although the objective function of this model capture the box prior,  similarity, and discriminability within different videos, as we can define a more efficient similarity mapping withing boxes in the sequence of frames in a video.

\subsection{Temporal Consistency In Frames of a Video}

As discussed earlier in this section, objects in consecutive frames in video data are less likely to change drastically in appearance, position, and size. This is a motivation to use a separate prior for frames or images in video case. Temporal consistency \cite{babenko2010robust,berclaz2011multiple,harestructured,pang2013finding,perez2002color,tang2012shifting,yilmaz2006object} is a powerful prior that is often leveraged in video tasks such as tracking \cite{joulin2014efficient}. In this approach, in consecutive frames, boxes with great difference in size and position should be unlikely to be selected together. To this end, a simple temporal similarity measure is defined between two boxes $b_i$ and $b_j$ from consecutive frames with:
\begin{align}\label{localization:formulation:temporal}
	s_{\text{temporal}}(b_i, b_j) := \exp\cbr{-\norm{b_i^{\text{center}} - b_j^{\text{center}}}_2 - \norm{\frac{\abs{b_i^{\text{area}} - b_j^{\text{area}}}}{\max(b_i^{\text{area}} , b_j^{\text{area}})}}_2}.
\end{align}

\begin{figure}[!ht]
	\centering
	\includegraphics[scale=.28]{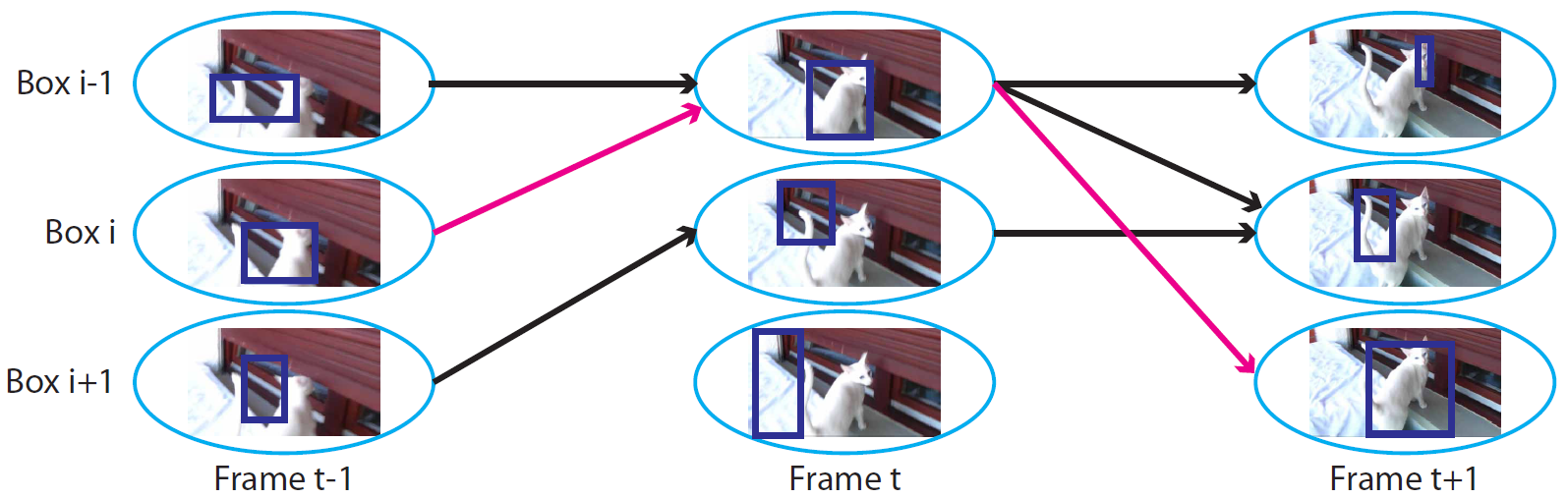}
	\caption{ Nodes (blue circles) represent candidate boxes in each frame, and the directed edges between nodes are weighted by a temporal similarity metric (e.g. \ref{localization:formulation:temporal}) that measures the similarity in size and position of boxes. To reduce the dimension of the graph, edges with low similarity are removed to limit the number of possible paths through the graph from the first to the last frame. The magneta edges represent the optimal path in this example. Image is from \cite{joulin2014efficient}.}
	\label{fig:localization:frames1}
\end{figure}

A few comments comes in place about the prior defines  in \eqref{localization:formulation:temporal}. First, $b_i^{\text{area}}$ is the vector of the pixel area of box $b_i$ and $b_i^{\text{center}}$ are the vectors of the center coordinates of box $b_i$, normalized by the width and height of the frame. Second, the metric defined in \eqref{localization:formulation:temporal} is a similarity metric that is defined between all pairs of boxes in adjacent frames. From this metric we can define a weighted graph $\mathcal{G}_i$ for video $\mathcal{V}_i$ for $i = 1,2, \dots, n$ with nodes being the boxes in each frame and edges connecting boxes in consecutive frames and weights of edges defined as temporal similarity in \eqref{localization:formulation:temporal}. Figure \ref{fig:localization:frames1} is a graphical representation of graph $\mathcal{G}_i$. For small values of similarity measure with some threshold we disconnect the nodes and remove the edge. Finally, as long as we can create a weighted graph with boxes, any similarity measure other than the temporal consistency in \eqref{localization:formulation:temporal} can be used to weight the edges between two boxes, which makes the temporal framework pretty flexible. 

Let us define
\begin{equation}\label{localization:video:simialrity}
	S_t(i,j) = \left\{
		\begin{array}{ll}
			s_{\text{temporal}}(b_i, b_j) & \text{ if frames $i$ and $j$ are adjacent}\\
			0 & \text{ otherwise}		
		\end{array}
		\right.
\end{equation}
to be the similarity matrix define by the temporal similarity measure, where $b_i$ and $b_j$ are any two boxes in the set of all boxes in $\mathcal{V}$. Similar to our approach to obtain \eqref{localization:laplacian}, with $S_t$ we can compute the normalized Laplacian 
\begin{align}\label{localization:video:laplacian}
	U =  I_{n_b^v} - D^{-1/2}S_tD^{-1/2},
\end{align}
where $D$ is the diagonal matrix composed of the row sums of $S_t$. This matrix encourages us to select boxes that are similar based on the temporal similarity metric \eqref{localization:formulation:temporal}.

\subsection{Video Model Formulation}

As we discussed above, temporal similarity suggests a weighted graph $\mathcal{G}_i$ for video $\mathcal{V}_i$ for $i=1,2,\dots,n$. In fact, a valid path in $\mathcal{G}_i$ from the the first to the last frame in $\mathcal{V}_i$ corresponds to feasible boxes chosen in each frame of $\mathcal{V}_i$. This motivates us to define a binary variable to be on when there is an edge between any two nodes in $\mathcal{G}_i$ and off otherwise. In better words, we define the binary variable $y_{i,j,k}$ for video $i$ and boxes $b_j$ and $b_k$ in $\mathcal{V}_i$ as
\begin{equation}\label{localization:video:yab}
	y_{i,j,k} = \left\{
		\begin{array}{ll}
			1 & \text{if boxes $b_j$ and $b_k$ contain the common object}\\
			0 & \text{otherwise.}
		\end{array}
	\right.
\end{equation}
In fact, variable $y_{i,j,k}$ corresponds to the existence of edge between boxes $b_j$ and $b_k$ in $\mathcal{V}_i$. Also, we define the binary variable $z_{i,j,k}$ to be 1 if the box $b_k$ in frame $j$ of video $i$ contains the common object, and 0 otherwise. A type of constraint that we need to consider here is the fact that there might exist an edge between boxes $b_j$ and $b_k$ only if they are boxes in two consecutive frames. Then, for a typical box $b_k$ in frame $j$ of video $\mathcal{V}_i$, we define index sets $p(k_j)$ and $c(k_j)$ to be the set of indices of parents and children boxes in frames $j+1$ and $j-1$, respectively, that are connected to $b_k$ in frame $j$ in the graph $\mathcal{G}_i$. Therefore, a required set of constraints for localization in video case are defines by:
\begin{align}\label{localization:video:cons1}
	z_{i,j,k} = \sum_{l\in p(k_j)} y_{i,l,k_j} = \sum_{l\in c(k_j)}y_{i,k_j,l},\qquad i = 1,\dots, n,\ j=1,\dots,l_i,\ k=1,\dots,m.
\end{align}

The other set of constraints, which are quite similar to the image co-localization case, are the set of constraints restricting each frame of each video to has only one box that contains the common object. These constraints are defined by:
\begin{align}\label{localization:video:cons2}
	\sum_{k = 1}^{m} z_{i,j,k} = 1,\qquad i=1,2,\dots,n,\ j = 1,2,\dots, l_i.
\end{align}

Finally, we define the vectors of variables 
\begin{align*}
    \vec{z} = (z_{1,1,1},z_{1,1,2}, \dots, z_{i,j,k}, \dots, z_{n,l_n,m})^T\in \{0,1\}^{n_b^v}
\end{align*}
where $n_b^v = m\sum_{i=1}^{n}l_i$. Then if we combine the temporal terms defined by \eqref{localization:video:laplacian} with the terms in the objective function of the original image model \eqref{localization:formualtion:img_problem}, then with constraint defines in \eqref{localization:video:cons1} and \eqref{localization:video:cons2}, we obtain the following optimization formulation to select the box containing the common object in each frame of video:
\begin{align}\label{localization:video:problem}
	\begin{aligned}
		\min_{\vec{z},\ y}&\qquad \vec{z}^T(L+\mu A + \mu_t U)\vec{z} - \lambda\; \vec{z}^T\log(\vec{m})\\
		\text{s.t.}&\qquad \sum_{k = 1}^{m} z_{i,j,k} = 1,\qquad i=1,2,\dots,n,\ j = 1,2,\dots, l_i,\\
		  	        &\qquad z_{i,j,k} = \sum_{l\in p(k_j)} y_{i,l,k_j} = \sum_{l\in c(k_j)}y_{i,k_j,l}\\
		  	        &\hspace{2cm} i = 1,\dots, n,\ j=1,\dots,l_i,\ k_j=1,\dots,m,\\
		  	        &\qquad y_{i,s,t}\in\{0,1\},\quad i = 1,\dots,n,\ s,t = 1,\dots,m\\
		  	        &\qquad \vec{z}=(z_{1,1,1},z_{1,1,2}, \dots, z_{i,j,k}, \dots, z_{n,l_n,m})^T \in \{0,1\}^{n_b^v},
	\end{aligned}
\end{align}
where $\mu_t$ is the trade-off weight for the temporal Laplacian matrix. Note that with the new objective function in problem \eqref{localization:video:problem} the extra constraint \eqref{localization:video:cons1} in video case is necessary  and without that the temporal Laplacian matrix would lead the solution to an invalid path. This formulation allows us to incorporate temporal consistency into the image model.

\section{Optimization }
The formulation \eqref{localization:formualtion:img_problem} obtained to find the best box in each image of the set of the given images is a standard binary constrained quadratic problem. The only issue that makes this problem a non-convex problem are the binary constraints. Relaxing these constraints to the continuous linear constraints lead the problem to the convex optimization problem and can be solved efficiently using standard methods. In fact, first order methods such as like Frank-Wolfe method that we discussed in previous chapters can handle the relaxed problem efficiently as they linearize the quadratic objective function and use a linear optimization oracle in each iteration. 

Denoting the feasible region of the problem \eqref{localization:video:problem} by $\mathcal{P}$, we can follow a similar approach for this problem as we did for \eqref{localization:formualtion:img_problem}. We can relax the discrete non-convex set $\mathcal{P}$ into the convex hull, or the integer hull for this specific case, conv($\mathcal{P}$). Although standard algorithms such as interior point methods can be applied to solve this problem, but as the number of videos increases to hundreds and the dimension of the problem increases exponentially, such problems with complexity of $\mathcal{O}(N^3)$ with number of boxes, would perform very weakly. Similarly, for the relaxation of the video problem we will show in our implementations section that suggested first order methods perform efficiently. We will also propose a first order method later in this chapter and will show that it performs better than other first order methods that have been applied to this problem.

Note that, the constraints defining the set $\mathcal{P}$ are separable in each video. In fact, for each video, these constraints are equivalent to the constraints of the shortest-path problem. This implies that the linear optimization step appears in each iteration of the first order methods are actually shortest-path problems that can be solved efficiently using dynamic programming. 

Recall that Frank-Wolfe algorithm is a first order method that in each of its iteration updates the new point toward a direction by calling a linear optimization oracle. This objective function of this linear optimization is in fact a linear approximation of the objective function of \eqref{localization:formualtion:img_problem}, and \eqref{localization:video:problem}. Frank-Wolfe algorithm specifically results in a simple linearizations with integer solution for the image and video co-localization optimization problems. For the image model, the linearlized cost function is separable for each image, and we can efficiently find the best integer solution with some threshold for this problem. For the video model also, the cost function and the constraints are separable for each video and optimizing the linearized function over the feasible region results in the shortest-path problem for each video. 

In the following section we will propose an algorithm that can be applied on image and video co-localization optimization problems efficiently and we finally compare the performance of the proposed algorithm to the algorithms that are applied to these problems.
 
\section{Proposed Algorithms}

Conditional Gradient Sliding (CGS) algorithm \cite{CGS:Lan}, is a first order projection free method for solving convex optimization problems in which the feasible region is a convex and compact set. The major advantage of the CGS algorithm is that it skips gradient evaluation from time to time and uses the same information within some inner iterations. This property of the CGS algorithm becomes helpful when the dimension of the problem as size of the variable is relatively large and computations become more and more expensive. 

As showed in previous chapters, CGS algorithm and its proposed variant, Conditional Gradient Sliding with Linesearch (CGS-ls) perform very well in many practical instances. Although the CGS and CGS-ls algorithms out-perform the Frank-Wolfe (FW) algorithm many cases, the variants of FW, such as Away-steps FW or Pairwise FW \cite{joulin2014efficient} converge faster to the optimal value than CGS for the image and video co-localization problem as we will show this in numerical experiments later in this chapter. 

Motivated from the CGS algorithm and also Away-steps and pairwise FW methods, we propose an algorithms called Away-Steps Conditional Gradient Sliding (ACGS) and Pairwise Conditional Gradient Sliding (PCGS) that perform very well for image and video co-localization problems. ACGS and PCGS methods have iterations of the CGS method but the direction to update the new point in each iteration is motivated from the away steps and pairwise steps in the Away-steps and Pairwise FW. We will also show that the ACGS and PCGS out-perform all of the variants of the FW applied to the image and Video co-localization problem.

\subsection{Away-Steps and Pairwise Conditional Gradient Sliding}

The basic scheme of the ACGS and PCGS methods is obtained by performing a new search direction in CGS method, if the new direction leads the algorithm to smaller Wolfe gap. Also, similar to the CGS algorithm, the classical FW method (as $\mathcal{FW}$ procedure)  is incorporated in this algorithm to solve the projection subproblems in the accelerated gradient (AG) with some approximations. The ACGS and PCGS algorithms are described as in \ref{Alg:ACGS} and \ref{Alg:PCGS}.

\begin{algorithm}[!ht]
	\caption{\label{Alg:ACGS} The Away-Steps Conditional Gradient Sliding Algorithm - Outer Iterations}
	\begin{algorithmic}
		\State Initial point $x_0\in \mathcal{A}$ and iteration limit $N$.\\
		\vspace{-.1cm} Let $\beta_k\in\R_{++}^n,\; \g_1 =1, \mathcal{S}^{(0)}:=\{x_0\}$, and $\eta_k\in\R_+,\; k=1,3\cdots$, be given and set $y_0=x_0$.
		\For {$k=1,\ldots,N$}
		\begin{align}
			&z\k = y\km + \g\k( x\km-y\km)\label{Alg:ACGS:z}\\
			&\nonumber\\
			&x_k = \mathcal{FW}(f'(z\k),x\km,\beta_k,\eta_k), \label{Alg:ACGS:x}\\
			&\nonumber\\
			&d_k^{\text{CGS}} = x\k - y\km ,\\
			&v_k = \argmax_{v\in \mathcal{S}^{(t)}} \ang{f'(y\km), v},\label{Alg:ACGS:v}\\
			&d_k^{\text{away}} = y\km - v\k\label{Alg:ACGS:away-dir}\\
			&\nonumber\\
			&\textbf{if } \ang{-f'(y\km), d_k^{\text{CGS}}}\le \epsilon\ \textbf{ then return } y\km\label{Alg:ACGS:stop}\\
			&\textbf{if } \ang{-f'(y\km), d_k^{\text{CGS}}} \ge \ang{-f'(y\km), d_k^{\text{away}}}, \textbf{ then }\label{Alg:away:start}\\
			&\qquad d\k := d_k^{\text{CGS}}\\
			&\qquad \g_{\text{max}}:= 1 \label{Alg:ACGS:gamma-max1}\\
			&\textbf{else }\\
			&\qquad d_k := d_k^{\text{away}}\\
			&\qquad \g_{\text{max}} = \alpha_{v_k}/(1-\alpha_{v_k})\label{Alg:ACGS:gamma-max2}\\
			&\textbf{end if}\label{Alg:away:end}\\
			&\g\kp \in \argmin_{\g\in[0,\g_{\text{max}}]} f(x\k + \g d_k) \label{Alg:ACGS:step-size}\\
			&y\k = y\km +\g\kp d\k \label{Alg:ACGS:s3}\\
			&\mathcal{S}^{(k)}:= \{v\in \mathcal{A};\ \alpha_v^{(k)}>0\}
		\end{align}
		\EndFor
		% \noindent \textbf{procedure} $u^+= \mathcal{FW}(g,u,\beta,\eta)$
		% \begin{enumerate}
		% 	\item Set $u_1=u$ and $t=1$.
		% 	\item Let $w_t$ be the optimal solution for the subproblem of
		% 	\begin{align}
		% 		\label{Alg:ACGS:linApp}
		% 		V_{g,u,\beta}(u_t):=\max_{x\in \mathcal{X}}\ang{g+\beta (u_t-u),u_t-x}
		% 	\end{align}
		% 	\item If $V_{g,u,\beta}(u_t)\le \eta$, set $u^+=u_t$ and terminate the procedure.
		% 	\item Set $u_{t+1}=(1-\tilde{\alpha}_t)u_t+\tilde{\alpha}_t w_t$, with
		% 	\begin{align}
		% 		\label{Alg:ACGS:stepsize}
		% 		\tilde{\alpha}_t=\min\cbr{1,\frac{\ang{\beta (u-u_t)-g,w_t-u_t}}{\beta\norm{w_t-u_t}^2}}
		% 	\end{align}
		% 	\item Set $t\leftarrow t+1$ and go to step 2.
		% \end{enumerate}
		% \noindent \textbf{end procedure}
	\end{algorithmic}
\end{algorithm}

\begin{algorithm}[!ht]
	\caption{\label{Alg:ACGS:2} The Away-Steps Conditional Gradient Sliding Algorithm - FW Procedure}
	\begin{algorithmic}
		\State \textbf{procedure} $u^+= \mathcal{FW}(g,u,\beta,\eta)$
		\begin{enumerate}
			\item Set $u_1=u$ and $t=1$.
			\item Let $w_t$ be the optimal solution for the subproblem of
			\begin{align}
				\label{Alg:ACGS:linApp}
				V_{g,u,\beta}(u_t):=\max_{x\in \mathcal{X}}\ang{g+\beta (u_t-u),u_t-x}
			\end{align}
			\item If $V_{g,u,\beta}(u_t)\le \eta$, set $u^+=u_t$ and terminate the procedure.
			\item Set $u_{t+1}=(1-\tilde{\alpha}_t)u_t+\tilde{\alpha}_t w_t$, with
			\begin{align}
				\label{Alg:ACGS:stepsize}
				\tilde{\alpha}_t=\min\cbr{1,\frac{\ang{\beta (u-u_t)-g,w_t-u_t}}{\beta\norm{w_t-u_t}^2}}
			\end{align}
			\item Set $t\leftarrow t+1$ and go to step 2.
		\end{enumerate}
		\noindent \textbf{end procedure}
	\end{algorithmic}
\end{algorithm}

Note that the purpose of the proposed algorithm is to be applied to the image and video co-localization problems \eqref{localization:formualtion:img_problem} and \eqref{localization:video:problem}. The objective function in both problems, as discussed before, are convex functions, and the feasible region is a set of finite binary vectors called \emph{atoms} in $\R^d$ for some $d$. We denote this set by $\mathcal{A}$ and its convex hull conv($\mathcal{A}$) by $\mathcal{M}$. As $\mathcal{A}$ is finite, $\mathcal{M}$ is a polytope. 

The first difference between the AGCS(PCGS) and the CGS method is that we incorporate the set $\mathcal{S}^{(k)}$ of active atoms in the ACGS(PCGS) algorithm. This set keeps record of atoms (integer points) in $\mathcal{A}$ that are being used for the \emph{away} direction $d_K^{\text{away}}$ at each iteration such that the point $y\k$ at current iteration is the sum of corners in $\mathcal{S}^{(k)}$ reweighted by $\alpha^{(k)}$. This direction that is given in \eqref{Alg:ACGS:away-dir}, is defined by finding the atom $v_k$ in $\mathcal{S}^{(k)}$ that maximized the potential of descent given by $\ang{-f'(y\km), y\km - v\k}$. Note that obtaining $v\k$ in $\eqref{Alg:ACGS:v}$ is fundamentally easier as the linear optimization is over the $\mathcal{S}^{(k)}$, the active set of possibly small finite set of points. 

The second difference is in the way we update the step-size to update the new iteration point. As we observe in \eqref{Alg:ACGS:step-size} we incorporate a line-search method to obtain a step-size with maximum reduction in the objective toward a prespecified direction from the point at current iteration. With $\g_{\text{max}}$ defined in \eqref{Alg:ACGS:gamma-max1} and \eqref{Alg:ACGS:gamma-max2} as the maximum step-size for the line-search step the algorithm guarantees that the new iterates $y\k = y\km +\g_{\text{max}} d_k^{\text{away}}$ stays feasible in each iteration. Note that the parameter $\g_k$ in CGS algorithm is required to be set up in appropriate way to maintain the feasibility in each iteration. Such set ups are represented in \cite{CGS:Lan} as ${\g}\k = 3/(k+2)$ and ${\g}\k = 2/(k+1)$ and in fact, we can us these set ups for CGS steps in step \eqref{Alg:ACGS:gamma-max1} as the upper bound for $\gamma_k$ instead of 1 in line-search step \eqref{Alg:ACGS:step-size}. Also, it is easy to check that for the special case of the image and video co-localization problem in which the objective is a convex quadratic function $\g\k$ in step \eqref{Alg:ACGS:step-size} has the closed form
\begin{align}\label{localiztion:gamma:closed-form}
	\g\k = -\frac{d^T \nabla f(x)}{d^T Q d},
\end{align}
if $Q\succeq 0$ is the quadratic term in the objective. This value is projected to 0 or $\g_{\text{max}}$ if is outside of the range $[0, \g_{\text{max}}]$ for \eqref{Alg:ACGS:step-size} case.

\begin{algorithm}[!ht]
	\caption{\label{Alg:PCGS} The Pairwise Conditional Gradient Sliding Algorithm}
	\begin{algorithmic}
		\State Initial point $x_0\in \mathcal{A}$ and iteration limit $N$.\\
		\vspace{-.1cm} Let $\beta_k\in\R_{++}^n,\; \g_1 =1, \mathcal{S}^{(0)}:=\{x_0\}$, and $\eta_k\in\R_+,\; k=1,3\cdots$, be given and set $y_0=x_0$.
		\For {$k=1,\ldots,N$}
% 		\State Compute
		\begin{align}
			&\hspace{-3cm}z\k = y\km + \g\k( x\km-y\km)\label{Alg:PCGS:z}\\
% 			&\nonumber\\
			&\hspace{-3cm}x_k = \mathcal{FW}(f'(z\k),x\km,\beta_k,\eta_k), \label{Alg:PCGS:x}\\
% 			&\nonumber\\
			&\hspace{-3cm}d_k^{\text{CGS}} = x\k - y\km ,\\
			&\hspace{-3cm}v_k = \argmax_{v\in \mathcal{S}^{(t)}} \ang{f'(y\km), v},\label{Alg:PCGS:v}\\
			&\hspace{-3cm}\nonumber\\
			&\hspace{-3cm}\textbf{if } \ang{-f'(y\km), d_k^{\text{CGS}}}\le \epsilon\ \textbf{ then return } y\km\label{Alg:PCGS:stop}\\
	   % \end{align}
	   % \If {\begin{align}\ang{-f'(y\km),\ d_k^{\text{CGS}}}\le \epsilon\label{Alg:PCGS:stop}\end{align}}{\textbf{ return} $y\km$}
	   % \begin{align}
	        &\hspace{-3cm}\nonumber\\
			&\hspace{-3cm} d\k^{\text{PCGS}} = x\k - v\k,\label{Alg:PCGS:start}\\
			&\hspace{-3cm} \g_{\text{max}} = \alpha_{v_k},\label{Alg:PCGS:end}\\
			&\hspace{-3cm}\g\kp \in \argmin_{\g\in[0,\g_{\text{max}}]} f(x\k + \g d\k^{\text{PCGS}}) \label{Alg:PCGS:step-size}\\
			&\hspace{-3cm}y\k = y\km +\g\kp d\k \label{Alg:PCGS:s3}\\
			&\hspace{-3cm}\mathcal{S}^{(k)}:= \{v\in \mathcal{A};\ \alpha_v^{(k)}>0\}
		\end{align}
% 		\EndIf
		\EndFor
		% \noindent \textbf{procedure} $u^+= \mathcal{FW}(g,u,\beta,\eta)$
		% \begin{enumerate}
		% 	\item Set $u_1=u$ and $t=1$.
		% 	\item Let $w_t$ be the optimal solution for the subproblem of
		% 	\begin{align}
		% 		\label{Alg:PCGS:linApp}
		% 		V_{g,u,\beta}(u_t):=\max_{x\in \mathcal{X}}\ang{g+\beta (u_t-u),u_t-x}
		% 	\end{align}
		% 	\item If $V_{g,u,\beta}(u_t)\le \eta$, set $u^+=u_t$ and terminate the procedure.
		% 	\item Set $u_{t+1}=(1-\tilde{\alpha}_t)u_t+\tilde{\alpha}_t w_t$, with
		% 	\begin{align}
		% 		\label{Alg:PCGS:stepsize}
		% 		\tilde{\alpha}_t=\min\cbr{1,\frac{\ang{\beta (u-u_t)-g,w_t-u_t}}{\beta\norm{w_t-u_t}^2}}
		% 	\end{align}
		% 	\item Set $t\leftarrow t+1$ and go to step 2.
		% \end{enumerate}
		% \noindent \textbf{end procedure}
	\end{algorithmic}
\end{algorithm}

Finally, we incorporate the Wolfe gap as an stopping criterion in the ACGS and PCGS algorithms. In fact, at steps \eqref{Alg:ACGS:stop} and \eqref{Alg:PCGS:stop}, the algorithms checks if they have reached the given threshold to stop before the preset max number of iterations $N$. As in classical FW, the Wolfe gap is an upper bound on the unknown suboptimality and from the convexity of the objective $f$ we have
\begin{align}\label{localization:ACGS:wolfe-gap}
	f(x_k) - f(x^{\star}) \le \ang{-f'(x\k), x^{\star}-y\km}\le \ang{-f'(x\k), x\k-y\km} \le \epsilon.
\end{align}

Note that for the image and video co-localization problem with binary decision variables in a CGS step we have
\begin{equation}
	\mathcal{S}^{(k+1)} = \left\{
	\begin{array}{ll}
		\{x_k\}						  & \text{ if } \g\k = 1\\
		\mathcal{S}^{(k)}\cup \{x\k\} & \text{ otherwise. }
	\end{array}
	\right.
\end{equation}
Also, for $v\in\mathcal{S}^{(k)}\setminus \{s_k\}$ we have
\begin{align}
	\alpha_{s_t}^{(k+1)}:=(1-\g\k)\alpha_{s_t}^{(k)} + \g\k\quad \text{ and }\quad \alpha_{v}^{(k+1)}:= (1-\g\k)\alpha_{v}^{(k)}.
\end{align}
On the other hand, for an away step we have
\begin{equation}
	\mathcal{S}^{(k+1)} = \left\{
	\begin{array}{ll}
		\mathcal{S}^{(k)}\setminus\{v\k\}	&	\text{ if } \g\k = \g_{\text{max}}\\
		\mathcal{S}^{(k)}				    &	\text{ otherwise. }
	\end{array}
	\right.
\end{equation}
This step is called a \emph{drop step}. Also, for $v\in\mathcal{S}^{(k)}\setminus \{v_k\}$ we have
\begin{align}
	\alpha_{v_t}^{(k+1)}:=(1+\g\k)\alpha_{v_t}^{(k)} + \g\k\quad \text{ and }\quad \alpha_{v}^{(k+1)}:= (1+\g\k)\alpha_{v}^{(k)}.
\end{align}

ACGS and PCGS algorithms are slightly different in the direction that they use to update the new point at each iteration. More precisely, steps \eqref{Alg:away:start} to \eqref{Alg:away:end} in Algorithm \ref{Alg:ACGS} are replaced with steps \eqref{Alg:PCGS:start} and \eqref{Alg:PCGS:end} in Algorithm \ref{Alg:PCGS}. Similar to the Paiwise FW, the idea here is to only move weight from the away atom $v\k$ to the CGS atom $x\k$ and keep all other $\alpha$ weight unchanged. In other words
\begin{align}\label{localization:PCGS:alpha}
	\alpha_{v_t}^{(k+1)}:=\alpha_{v_t}^{(k)} - \g\quad \text{ and }\quad \alpha_{x\k}^{(k+1)}:= \alpha_{s\k}^{(k)}+\g,
\end{align}
for some $\g\le \g_{\text{max}}:=\alpha_{v_t}^{(k)}$.

An important property of the formulation \eqref{localization:formualtion:img_problem} and \eqref{localization:video:problem} is that their constraints are separable for each image and video. This helps computation to be more efficient if we use parallel computing. This, however, is a property of any first-order method and practically it is very memory efficient. In addition, as a solution to the convex relaxation is not necessarily an integer solution optimal or feasible to the original problem, we need to come up with a solution as close as possible to the obtained relaxation optimum. In image and video co-localization case, the most natural way of finding such a solution is to solve 
\begin{align}\label{localization:rounding1}
	\min_{p\in\mathcal{P}}\quad \norm{p - y}_2^2,
\end{align}
where $\mathcal{P}$ is the feasible region of the original problem and $y$ is the solution to the relaxed problem. It is easy to check that the projection problem \eqref{localization:rounding1} is equivalent to 
\begin{align}\label{localization:rounding2}
	\max_{p\in\mathcal{P}}\quad \ang{p,y},
\end{align}
which for the video model is just a shortest path problem that can be solved efficiently using dynamic programming.

\section{Experimental Results}

In this section we experiment the proposed Algorithm \ref{Alg:ACGS} to the problems introduced in \eqref{localization:formualtion:img_problem} and \eqref{localization:video:problem} for image and video co-localization task. Recall that these problems are quadratic problems over the convex hull of paths in a network, the linear minimization oracle in first order methods is equivalent to find a shortest path in the network. We compare the performance of the proposed algorithm with the works in \cite{joulin2014efficient} and \cite{lacoste2015global} on FW algorithm and its variants for the similar problem. For this comparison we reuse the codes available and shared for \cite{lacoste2015global,joulin2014efficient,tang2014co} and the included dataset of airplanes consist of 660 variables. 

We begin this section by reviewing the performance of Away steps Frank-Wolfe (AFW) and its comparison to the solvers such as Gurobi and Mosek. These results are derived and shown in \cite{joulin2014efficient} and the goal in this section is to show how AFW outperforms other methods for our problem of interest. In \cite{lacoste2015global}, however, Joulin A., Tang K., and Fei-Fei L. showed that their proposed Pairwise Frank-Wolfe (PairFW) algorithm outperforms any other variants of FW in solving this problem. We will end this section by showing that our proposed ACGS algorithm performs better any first order methods that have been utilized to solve the video co-localization problem.

\subsection{FW v.s. Mosek and Gurobi}

Algorithm \ref{Alg:AFW-local} is a variant of FW algorithm proposed in \cite{joulin2014efficient} in which  the authors examined it on two datasets, the PASCAL VOC 2007 dataset \cite{everingham2010pascal} and the Youtube-Objects dataset \cite{prest2012learning}. This algorithm is in fact the AWF Algorithm introduced in \cite{joulin2014efficient} with some slight changes and some extra rounding steps. Also, the set $\mathcal{D}$ in this algorithm is conv$(\mathcal{P})$ the convex hull of the feasible region of problems \eqref{localization:formualtion:img_problem} or \eqref{localization:video:problem}. Their implementation of Algorithm \ref{Alg:AFW-local} was coded in MATLAB and they compare it to two standard Quadratic Programming (QP) solvers, Mosek and Gurobi on a single-core 2.66GHz Intel CPU with 6GB of RAM. In addition, they set $\mu=0.4$ for the image model and $\mu=0.6$ for the video model and $\mu_t=1.8$ and $\lambda = 0.1$, for both image and video models. They extracted 20 objectness boxes from each image and sample each video every 10 frames as there is little change frames in short amount time. 

\begin{algorithm}[!ht]
	\caption{\label{Alg:AFW-local} Frank-Wolfe Algorithm with Away Steps and Rounding \cite{joulin2014efficient}}
	\begin{algorithmic}
		\State Initialization $y_0\in \mathcal{D}, \epsilon >0,\ k=0,\ z=y_0, \mathcal{S}_0=\{y_0\},\ \alpha_0=\{1\}$.\\
		\textbf{while} duality-gap$(z)\ge \epsilon$ \textbf{do}
		\begin{align}
			&k \leftarrow k+1;\\
			&y\k \leftarrow \argmin_{y\in\mathcal{D}}\ang{y,\nabla f(z)} \text{ (FW direction);}\\
			&x\k \leftarrow \argmax_{y\in\mathcal{S}\km}\ang{y,\nabla f(z)}\text{ (away direction);}\\
			&\textbf{if } \ang{y\k - z, \nabla f(z)} \le \ang{z-x\k, \nabla f(z)} \text{ \textbf{then}}\\
			&\hspace{1cm} d\k = y\k - z;\\
			&\hspace{1cm} \gamma_{\text{max}} = 1;\\
			&\textbf{else}\\
			&\hspace{1cm} d\k = z - x\k;\\
			&\hspace{1cm} \gamma_{\text{max}} = \alpha_k(x\k);\\
			&\textbf{end}\\
			&\gamma\k = \min_{\g\in[0, \gamma_{\text{max}}]} f(z+\g d\k);\\
			&\mathcal{S}\k,\alpha\k \leftarrow update\_active\_set(d\k,\g\k);\\
			&\text{Update } z\leftarrow z+\g\k d\k;\\
			&\textbf{if } f(y\k)< f(y^{\star}) \textbf{ then}\\
			&\hspace{1cm} y^{\star} \leftarrow y\k \text{ (rounding 1)};\\
			&\textbf{end}
		\end{align}
		\textbf{end while}
		
		$y_r\leftarrow \argmax_{y\in\mathcal{D}}\ang{y,z}$;
		
		\textbf{if } $f(y_r)<f(y^{\star})$ \textbf{ then}
		
		\hspace{1cm} $y^{\star} \leftarrow y_r$ (combining rounding);
		
		\textbf{end}
	\end{algorithmic}
\end{algorithm}

The stopping criterion of Algorithm \ref{Alg:AFW-local} is based on the relative duality gap. This criterion, that is given in function duality-gap($z$) in the algorithm, is defined as $d = (f-g)/g$, where $f$ is the objective function and $g$ is its dual. In the implementation of this algorithm, authors consider two values $1e$ - 2 and $1e$ - 3 for the stopping threshold $\epsilon$.

\begin{figure}[!ht]
	\centering
	\includegraphics[scale=.32]{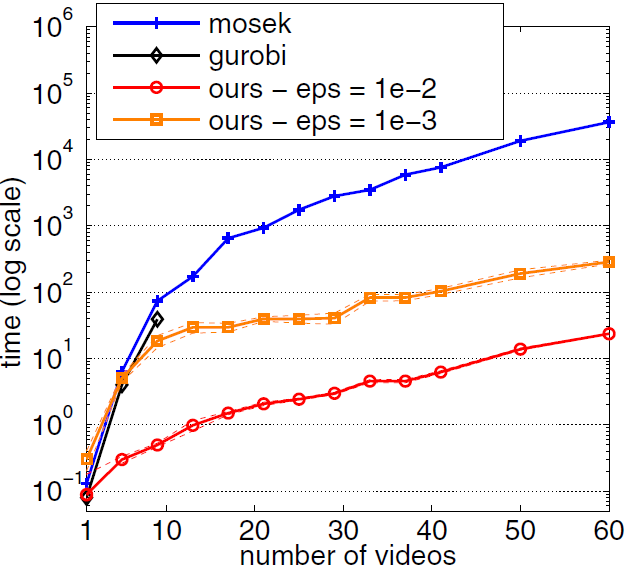}
	\hspace{1cm}
	\includegraphics[scale=.32]{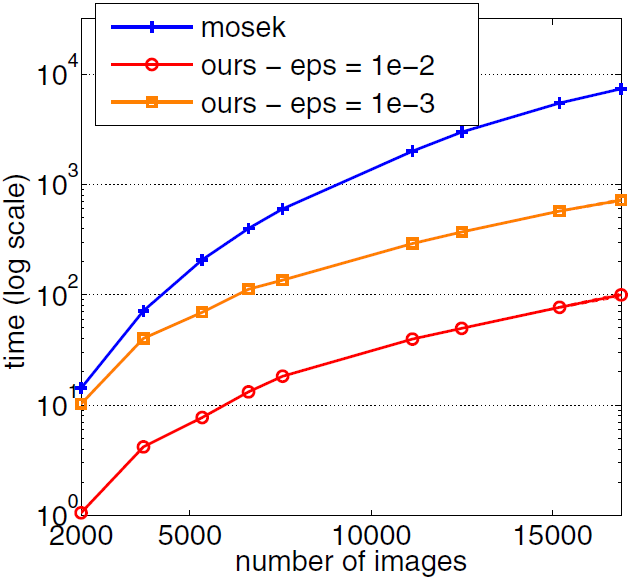}
	\caption{ "Ours" in the legend of these plots refers to the Algorithm \ref{Alg:AFW-local}. Note that for $\epsilon=1e-3$ Algorithm \ref{Alg:AFW-local} performs 100 times faster than standard solvers for more than 20 videos. Plots are from \cite{joulin2014efficient}.
	}
	\label{fig:localization:FW-loc}
\end{figure}

Figures \ref{fig:localization:FW-loc} presents some comparisons of the Algorithm \ref{Alg:AFW-local} as a variant of FW algorithm with QP solvers Mosek and Gurobi in logarithmic scale.  Indeed, this comparison is based on the CPU time performance of the algorithms depending on the number of images and videos, or in better words, the dimension of the decision variables. This time is the time that takes that algorithms reach a duality gap less than the threshold $\epsilon$. As we can observe from these plots, the variant of FW algorithm with away steps outperforms the standard QP solvers Mosek and Gurobi. 

The reason that we review and represent these comparisons directly from \cite{joulin2014efficient}local is that in our implementations in next section we will only compare our proposed algorithms to some other first order methods. These first order methods include the AWF algorithm that we already know from this section that it outperforms standard QP solvers.

The PASCAL Visual Object Classes 2007 dataset \cite{everingham2010pascal} provides standardized image data of 20 objects for object classes recognition along with annotations for images and bounding box and object class label for each object. Challenges and competitions have been used to recognize objects from a number of visual object classes in realistic scenes. The YouTube-Objects dataset \cite{prest2012learning} consists of YouTube videos collected for 10 classes from PASCAL \cite{everingham2010pascal}: "aeroplane", "bird", "boat", "car", "cat", "cow", "dog", "horse", "motorbike", and "train". Although authors in \cite{joulin2014efficient} did the study on multiple objects of this dataset, in our implementations our focus will be on the "aeroplane" object class.

% \begin{figure}[!ht]
% 	\centering
% 	\includegraphics[scale=.3503]{imgs/video-solution.png}
% 	\caption{ Example co-localization results on YouTube-Objects for the video model \ref{localization:video:problem} with optimal green boxes and the image model \ref{localization:formualtion:img_problem} with optimal red boxes. Each column corresponds to a different class, and consists of frame samples from a single video. Image is from \cite{joulin2014efficient}.
% 	}
% 	\label{fig:localization:video-solution}
% \end{figure}

\subsection{Implementations}

Knowing that AFW Algorithm \ref{Alg:AFW-local} outperforms the standard QP solvers Mosek and Gurobi from the works in \cite{joulin2014efficient}, in this section we compare our proposed variants of the CGS algorithm, the ACGS Algorithm \ref{Alg:ACGS} and the PCGS Algorithm \ref{Alg:PCGS} to some other first order methods, including the AFW method. More precisely, we will compare the performance of our algorithms to all of the variants of the FW namely, the FW, the FW Algorithm with away steps (AFW), and the pairwise FW Algorithm as discussed in \cite{joulin2014efficient}. We also compare our algorithms to the original CGS Algorithm \cite{CGS:Lan}. These comparisons include the duality gap, CPU time, and objective function value versus the iterations. 

The implementations are over the YouTube Objects dataset \cite{everingham2010pascal} explained in previous section, and specifically its "aeroplane" class. We obtain the dataset for this class and also the codes for AFW and Pairwise FW algorithms available in the repositories for \cite{joulin2014efficient,lacoste2015global,tang2014co}. We only consider the task of video co-localization with the problem formulation defined in \eqref{localization:video:problem} for this implementation. All algorithms are coded in MATLAB and run on a computer with Interl Core i5-6500 CPU 3.2 GHz processor with 16 GB of RAM.

In our implementations, we set all algorithms to stop either after the maximum number of iterations or after reaching the Wolfe duality gap threshold. We set the threshold to $\epsilon=1e-5$ and the max number of iterations to 2000 iterations. All of the parameters exist in \eqref{localization:video:problem} are set the same as in \cite{lacoste2015global} for consistency in the comparison.

\begin{figure}[!ht]
	\centering
	\includegraphics[scale=.36]{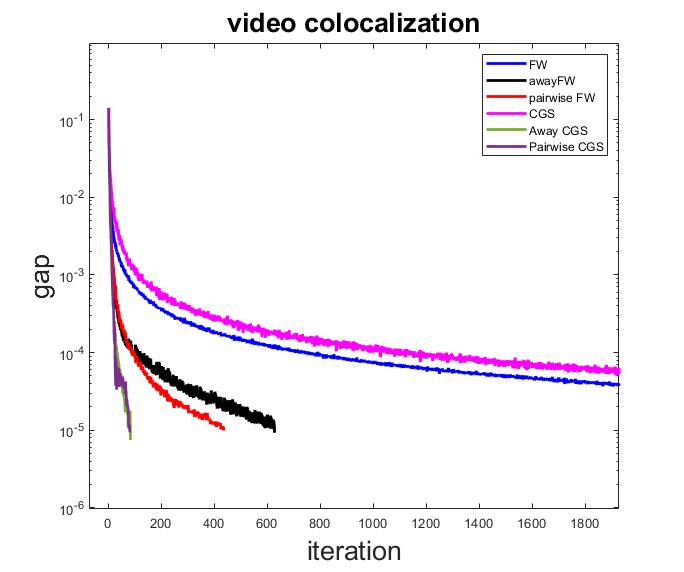}
	% \hspace{1cm}
	% \includegraphics[scale=.45]{imgs/loc-objVal.png}
	\caption{On the left we observe the difference in gap reduction and on the right the objective value improvement, both with iteration increments. Here $\epsilon=1e-5$ and max number of iteration is 2000.
	}
	\label{fig:localization:gap-objVal-loc}
\end{figure}

Note that both original versions of FW and CGS algorithms do not reach the desired duality gap before the preset 2000 max number of iterations. Also, the AFW algorithm takes 628 iterations, the Pairwise FW takes 436 iterations, the ACGS takes 84 iterations, and PCGS takes 82 iterations to reach the threshold for the duality gap. 

\begin{figure}[!ht]
	\centering
	\includegraphics[scale=.6]{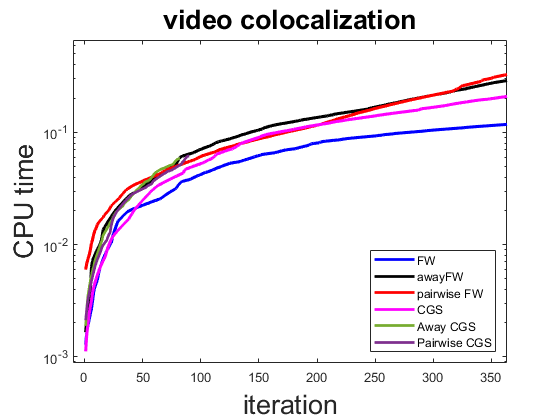}
	\caption{The difference in CPU time of the algorithms versus the iteration increments with $\epsilon=1e-5$ and max number of iteration is 2000.
	}
	\label{fig:localization:CPU-loc}
\end{figure}

As we observe in Figure \ref{fig:localization:gap-objVal-loc} both proposed variants of CGS algorithm, the ACGS and PCGS algorithms outperform the FW algorithms and its variants as well as the original CGS algorithm. The performance of the algorithms in terms of the CPU time versus iterations increments also is represented in Figure \ref{fig:localization:CPU-loc}. As we observe in this figure the CPU time per iteration of AFW and ACGS and PCGS are quite similar, although the ACGS and PCGS algorithms reach the gap much earlier than the AFW algorithm.

 In addition, while FW algorithm requires one linear optimization oracle per iteration, its CPU time per iteration is not significantly better than the other algorithms. Also, note that out of 84 iteration of the ACGS algorithm, it chooses the away direction in 34 iteration which improves the performance of CGS (with more than 2000 iterations) for this problem significantly. 

Finally, authors in \cite{lacoste2015global} proved, for the first time, the global linear convergence of the variants of FW algorithms, AFW and Pairwise FW, under strong convexity of the objective. One potential research work related to the current chapter is figure out the convergence of the proposed algorithms \ref{Alg:ACGS} and \ref{Alg:PCGS}.



\begin{thebibliography}{}
%
\bibitem{CGS:Lan}
Lan, Guanghui, and Yi Zhou. "Conditional gradient sliding for convex optimization." SIAM Journal on Optimization 26.2 (2016): 1379-1409

\bibitem{Nesterov}
Nesterov, Y.: Introductory lectures on convex optimization: A basic course, vol. 87. Springer Science \& Business Media (2013)

\bibitem{joulin2014efficient}
Joulin, A., Tang, K., Fei-Fei, L.: Efficient image and video co-localization with frank-wolfe algorithm. In: European Conference on Computer Vision, pp. 253–268. Springer (2014)

\bibitem{tang2014co}
Tang, K., Joulin, A., Li, L.J., Fei-Fei, L.: Co-localization in real-world images. In: Proceedings of the IEEE conference on computer vision and pattern recognition, pp. 1464–1471 (2014)

\bibitem{alexe2012measuring}
Alexe, B., Deselaers, T., Ferrari, V.: Measuring the objectness of image windows. IEEE trans-actions on pattern analysis and machine intelligence 34(11), 2189–2202 (2012)

\bibitem{boykov2001fast}
Boykov, Y., Veksler, O., Zabih, R.: Fast approximate energy minimization via graph cuts. IEEE Transactions on pattern analysis and machine intelligence 23(11), 1222–1239 (2001)

\bibitem{delong2012minimizing}
Delong, A., Gorelick, L., Veksler, O., Boykov, Y.: Minimizing energies with hierarchical costs. International journal of computer vision 100(1), 38–58 (2012)

\bibitem{delong2012fast}
Delong, A., Osokin, A., Isack, H.N., Boykov, Y.: Fast approximate energy minimization with label costs. International journal of computer vision 96(1), 1–27 (2012)

\bibitem{lowe2004distinctive}
Lowe, D.G.: Distinctive image features from scale-invariant keypoints. International journal of computer vision 60(2), 91–110 (2004)

\bibitem{perazzi2012saliency}
Perazzi, F., Krauhenbuhl., Pritch, Y., Hornung, A.: Saliency filters: Contrast based filtering for salient region detection. In: 2012 IEEE conference on computer vision and pattern recognition, pp. 733–740. IEEE (2012)

\bibitem{shi2000normalized}
Shi, J., Malik, J.: Normalized cuts and image segmentation. IEEE Transactions on pattern analysis and machine intelligence 22(8), 888–905 (2000)

\bibitem{belkin2003laplacian}
Belkin, M., Niyogi, P.: Laplacian eigenmaps for dimensionality reduction and data representation. Neural computation 15(6), 1373–1396 (2003)

\bibitem{bach2007diffrac}
Bach, F., Harchaoui, Z.: Diffrac: a discriminative and flexible framework for clustering. Advances in Neural Information Processing Systems 20 (2007)

\bibitem{xu2004maximum}
Xu, L., Neufeld, J., Larson, B., Schuurmans, D.: Maximum margin clustering. Advances in neural information processing systems 17 (2004)

\bibitem{joulin2010discriminative}
Joulin, A., Bach, F., Ponce, J.: Discriminative clustering for image co-segmentation. In: 2010 IEEE Computer Society Conference on Computer Vision and Pattern Recognition, pp. 1943–1950. IEEE (2010)

\bibitem{hastie2009elements}
Hastie, T., Tibshirani, R., Friedman, J.: The Elements of Statistical Learning: Data Mining, Inference, and Prediction, Second Edition. Springer Series in Statistics. Springer (2009)

\bibitem{babenko2010robust}
Babenko, B., Yang, M.H., Belongie, S.: Robust object tracking with online multiple instance learning. IEEE transactions on pattern analysis and machine intelligence 33(8), 1619–1632 (2010)

\bibitem{berclaz2011multiple}
Berclaz, J., Fleuret, F., Turetken, E., Fua, P.: Multiple object tracking using k-shortest paths optimization. IEEE transactions on pattern analysis and machine intelligence 33(9), 1806–1819 (2011)

\bibitem{yilmaz2006object}
Yilmaz, A., Javed, O., Shah, M.: Object tracking: A survey. Acm computing surveys (CSUR) 38(4), 13–es (2006)

\bibitem{tang2012shifting}
Tang, K., Ramanathan, V., Fei-Fei, L., Koller, D.: Shifting weights: Adapting object detectors from image to video. Advances in Neural Information Processing Systems 25 (2012)

\bibitem{perez2002color}
Perez, P., Hue, C., Vermaak, J., Gangnet, M.: Color-based probabilistic tracking. In: European Conference on Computer Vision, pp. 661–675. Springer (2002)

\bibitem{pang2013finding}
Pang, Y., Ling, H.: Finding the best from the second bests-inhibiting subjective bias in evaluation of visual tracking algorithms. In: Proceedings of the IEEE International Conference on omputer Vision, pp. 2784–2791 (2013)

\bibitem{harestructured}
Hare, S., Saffari, A., Torr, P., Struck, S.: Structured output tracking with kernels. In: IEEE International Conference on Computer Vision. IEEE, pp. 263–27

\bibitem{lacoste2015global}
Lacoste-Julien, S., Jaggi, M.: On the global linear convergence of frank-wolfe optimization variants. Advances in neural information processing systems 28 (2015)

\bibitem{everingham2010pascal}
Everingham, M., Van Gool, L., Williams, C.K., Winn, J., Zisserman, A.: The pascal visual
object classes (voc) challenge. International journal of computer vision 88(2), 303–338 (2010)

\bibitem{prest2012learning}
Prest, A., Leistner, C., Civera, J., Schmid, C., Ferrari, V.: Learning object class detectors from weakly annotated video. In: 2012 IEEE Conference on computer vision and pattern recognition,
pp. 3282–3289. IEEE (2012)


\end{thebibliography}
\end{document}